\documentclass{article}
\usepackage{amsmath,graphicx,hyperref}
\usepackage[preprint]{spconf}
\usepackage[doipre={DOI: }]{uri}

\usepackage{cite}
\usepackage{amsmath,amssymb,amsfonts}
\usepackage{enumitem}
\usepackage{algorithmic}
\usepackage{graphicx}
\usepackage{textcomp}
\usepackage{xcolor}
\usepackage{todonotes}
\usepackage{dirtytalk}
\usepackage{siunitx}
\usepackage{url}
\usepackage{hyperref}
\usepackage{subcaption}
\usepackage{paralist}
\usepackage[nice]{nicefrac}
\usepackage{orcidlink}
\usepackage{tabularx}
\usepackage{booktabs}
\usepackage{csvsimple}

\usepackage{multirow}
\usepackage{multicol}
\usepackage[ruled,vlined]{algorithm2e}
\usepackage{adjustbox}
\usepackage{tikz,pgfplots,pgfkeys}
\pgfplotsset{compat=1.17}
\newcommand\inputpgf[2]{{
		\let\pgfimageWithoutPath\pgfimage
		\renewcommand{\pgfimage}[2][]{\pgfimageWithoutPath[##1]{#1/##2}}
		\input{#1/#2}
}}
\usepackage{import}
\graphicspath{{./figs/svg}}

\DeclareUnicodeCharacter{2212}{-}

\usepackage{bm}
\usepackage{mathtools}

\DeclareUnicodeCharacter{03C4}{$\tau$}

\copyrightnotice{%
    \footnotesize
    \makebox[\textwidth][l]{%
        \parbox[t]{\textwidth}{%
          \raggedright
          \copyright\ 2026 IEEE. Personal use of this material is permitted.
          Permission from IEEE must be obtained for all other uses, in any current
          or future media, including reprinting/republishing this material for
          advertising or promotional purposes, creating new collective works,
          for resale or redistribution to servers or lists, or reuse of any
          copyrighted component of this work in other works.
          \\
          \doi{10.1109/ICASSP55912.2026.11462172}
        }%
    }%
}

\title{Cooperative Multi-Agent Reinforcement Learning for Adaptive Aggregation in Semi-Supervised Federated Learning with Non-IID Data}
\name{
Rene Glitza, 
Luca Becker and
Rainer Martin
}
\address{
Institute of Communication Acoustics,
Ruhr-Universit\"at Bochum,
Bochum, Germany
}

\begin{document}
\ninept
\maketitle
\begin{abstract}
Federated Learning (FL) enables distributed training of machine learning models while preserving data privacy. However, FL struggles with heterogeneous, non-IID client data distributions, resulting in sub-optimal and biased global models. In this paper, we propose pFedMARL, a novel approach leveraging Multi-Agent Reinforcement Learning (MARL) with Twin Delayed Deep Deterministic Policy Gradient (TD3) to dynamically adapt aggregation strategies in FL settings. Our method employs a server-side agent adjusting client contributions to optimize global model robustness and client-side agents balancing global and local updates to personalize models effectively without pre-training. We demonstrate superior performance of pFedMARL for training a semi-supervised audio spectrogram transformer, matching or outperforming FedAvg, Ditto, and local training approaches across multiple non-IID scenarios and in the presence of  adversarial clients. Our results indicate that pFedMARL actively improves accuracy, robustness, and fairness, making it suitable for real-world deployments.
\end{abstract}

\section{Introduction}
\label{sec:intro}
The rise of IoT devices with AI accelerators
has enabled the deployment of deep neural networks (DNNs) directly on consumer hardware. Running DNNs locally reduces latency and addresses growing privacy concerns, especially under data protection laws such as the EU GDPR \cite{GDPR}. 
However, to achieve high performance in real-world applications, access to diverse and authentic data remains critical.

Federated Learning (FL) \cite{mcmahan_CommunicationEfficientLearningDeep_2017} offers a promising solution by allowing edge devices to collaboratively train models without sharing raw data. Clients train locally and send model updates to a central server, which aggregates them into a global model and redistributes this model to clients. While effective in principle, FL struggles with non-independent and non-identically distributed (non-IID) data. Due to diverse environments and usage patterns, the distributions of client data may vary significantly \cite{li_FederatedOptimizationHeterogeneous_2020, hsieh_NonIIDDataQuagmire_2020}.

To overcome these issues, prior work has explored client selection, data weighting \cite{ wu_FastConvergentFederatedLearning_2021}, aggregation adjustments \cite{karimireddySCAFFOLDStochasticControlled2020}, and clustering \cite{sattler_ClusteredFederatedLearning_2021}. Yet, standard FL methods like Federated Averaging (FedAvg) \cite{mcmahan_CommunicationEfficientLearningDeep_2017} often result in high performance on average but are inconsistent across individual clients. Therefore, personalized FL aims to jointly optimize global and local models \cite{li_DittoFairRobust_2021}, or embed shared knowledge into clients \cite{huang_ReinforcementLearningBasedLayerWise_2024}. Also, adversarial clients may send malicious model updates that threaten the robustness and performance of the global model.
Server-side Deep Reinforcement Learning (DRL) like in FedAA \cite{he_FedAAReinforcementLearning_2025} and FedDRL \cite{nguyen_FedDRLDeepReinforcement_2023} can be used to adaptively weigh client contributions, filtering out adversarial clients and improving convergence and fairness, but does not improve personalization.

In this work, we propose pFedMARL
-- \textbf{P}ersonalized \textbf{Fed}erated Learning with \textbf{M}ulti-\textbf{A}gent Off-Policy \textbf{R}einforcement \textbf{L}earning -- a novel multi-agent RL-based FL framework that simultaneously creates a robust global model and personalized client models tailored to heterogeneous environments.
A server-side DRL agent dynamically adjusts client aggregation weights to enhance global model robustness against non-IID data distribution and adversarial behaviors.
Each client deploys a DRL agent locally to personalize its model by balancing individual learning with global knowledge.
We employ Twin Delayed DDPG (TD3) \cite{fujimoto_AddressingFunctionApproximation_2018}, trained online without pretraining, for both server and client agents.
Finally, we evaluate our method on a semi-supervised audio spectrogram transformer pre-training, benchmarking against FedAvg, Ditto, local-only training, and a global oracle system. By training masked-patch reconstruction and patch-level classification simultaneously, the transformer model learns rich and transferable audio representations that are well-suited for fine-tuning on various downstream applications, such as anomaly detection and condition monitoring \cite{wilkinghoff_SelfSupervisedLearningAnomalous_2024}, sound event detection \cite{shul_CSTFormerTransformerChannelSpectroTemporal_2024}, and speaker verification \cite{ecapa_tdnn}, where both generalization and accuracy are essential.
Here, pFedMARL provides a flexible and scalable solution to FL under non-IID conditions, offering strong performance in global cooperation and local personalization.

\section{Relation to prior work}
\label{sec:prior}
We situate pFedMARL within research on non-IID federated aggregation with adversarial clients, personalized FL, and DRL.
\subsection{Federated Learning}
FL enables collaborative training of a global model by having a central server aggregate weight updates from multiple clients in an iterative fashion, each of which trains locally on its private data. A common aggregation strategy is FedAvg \cite{mcmahan_CommunicationEfficientLearningDeep_2017}, defined as
\begin{equation}
    \label{eq:agg_server}
    \bm{\theta}_g^{\left<\tau\right>} = \sum_{i = 1}^M a_{g,i} \bm{\theta}^{\left<\tau\right>}_{i} , \ \text{with} \ \bm{\theta}_i^{\langle \tau \rangle} = \bm{\theta}_i^{\langle \tau - 1 \rangle} + \Delta \bm{\theta}_i^{\langle \tau \rangle},
\end{equation}
where $\bm{\theta}_g^{\left<\tau\right>}$ and $\bm{\theta}_i^{\left<\tau\right>}$ represent server and client model weights of communication round $\tau$, respectively, and $a_{g,i}$ are the aggregation weights summing to 1. Although FedAvg works well in general, it struggles under non-IID data conditions \cite{hsieh_NonIIDDataQuagmire_2020}.


Ditto \cite{li_DittoFairRobust_2021} is a framework for personalized FL that attempts to enhance FL regarding fairness in non-IID scenarios and robustness against label poisoning, random-update, and model-replacement attacks. It does so by jointly training a global model and client-specific models via a global-regularized multi-task objective.
Each client $i$ learns a personalized model by solving
\begin{equation}
\min_{\bm{\theta}_i^{\langle \tau \rangle}}\; F_i(\bm{\theta}_i^{\langle \tau \rangle}) + \frac{\lambda}{2}\,\left\|\bm{\theta}_i^{\langle \tau \rangle} - \bm{\theta}_g^{\langle \tau \rangle}\right\|_2^2
\label{eq:ditto_obj}
\end{equation}
where $F_i$ is the local objective and $\lambda$ trades off local adaptation against proximity to the global solution. When $\lambda = 0$, Ditto simplifies to training local models, and with $\lambda \rightarrow \infty$, it reverts to the global model objective.
Ditto alternates per round between updating the global model with a standard FedAvg solver as in (\ref{eq:agg_server}) and taking additional local steps on the personalized objective using the current global weights as described in (\ref{eq:ditto_obj}).

\subsection{Challenging non-IID data distributions}
Following \cite{nguyen_FedDRLDeepReinforcement_2023, hsieh_NonIIDDataQuagmire_2020}, our study investigates non-IID data category combinations of quantity, label, and cluster skew to realistically address FL challenges.
\textbf{Quantity Skew (QS):} Clients hold different amounts of data, frequently observed in user-generated datasets.
\textbf{Label Skew (LS):} Clients 
have distinct label distributions arising naturally from user-specific environments.
\textbf{Cluster Skew (CS):} Clients form groups with similar within-cluster but different across-cluster label distributions. This scenario closely reflects real-world conditions, capturing inter-client correlations prevalent in healthcare, regional deployments, and social settings \cite{nguyen_FedDRLDeepReinforcement_2023}.

\subsection{Deep Reinforcement Learning}
\label{sec:td3}
\begin{figure}
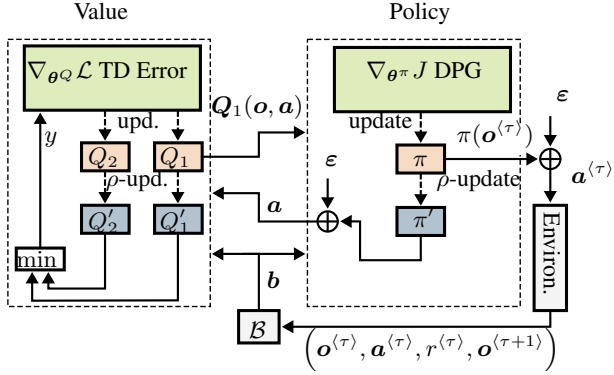

    \centering
    \vspace{0.7cm}
    \def\svgwidth{0.9\linewidth}
    \include{fig/system/td3}
    \vspace{-0.7cm}
    \caption{Illustration of the TD3 agent with an experience replay buffer $\mathcal{B}$, two main (red) and soft-updated target (blue) values networks 
    $Q_k$ and $Q'_k$, a main $\pi$ and target $\pi'$ policy network updated off-policy with noise added to target action $\bm{a}' + \bm{\varepsilon}$. $\pi(\bm{o}^{\left<\tau\right>})$ interacts based on observation $\bm{o}^{\left<\tau\right>}$ in FL round $\tau$
    with the environment for storing new trajectories $\bm{b}$.}
    \label{fig:agent}
\end{figure}

In Deep Reinforcement Learning (DRL), an \textit{agent} learns to interact with an \textit{environment} that has a \textit{state} $\bm{s}^{\left<\tau\right>} \in \mathcal{S}$ at timestamp $\tau$. The agent selects an \textit{action} $\bm{a}^{\left<\tau\right>} \in \mathcal{A}$ based on $\bm{s}^{\left<\tau\right>}$ to interact with the environment. Subsequently, it receives a next state $\bm{s}^{\left<\tau+1\right>}$, and a \textit{reward} $r^{\left<\tau\right>}$, representing the quality of the action.

We instantiate an actor–critic architecture with Twin Delayed Deep Deterministic Policy Gradient (TD3) \cite{fujimoto_AddressingFunctionApproximation_2018} as illustrated in Fig. \ref{fig:agent}. The actor is a $\bm{\theta}^{\pi}$-parameterized deterministic policy $\pi_{\bm{\theta}^{\pi}}(\bm{s})$ (abbreviated as  $\pi(\bm{s})$) that maximizes the expected cumulative discounted \textit{return} for an infinite horizon $\left(J\left(\bm{\theta}^{\pi}\right)\right)$ using the deterministic policy algorithm \cite{lillicrap_ContinuousControlDeep_2019}.

Two critics $Q(\bm{s},\bm{a})$ with parameters $\bm{\theta}^Q_k$, $k\in{1,2}$, are trained with temporal difference (TD) with clipped smoothing noise $\bm{\varepsilon}\sim\mathcal{N}(\bm{0},\sigma^2\bm{I})$ on the target action during training to estimate the expected return 
$\left(\mathcal{L}\left(\bm{\theta}^Q_k\right)\right)$.

Target networks $Q_{1,2}'$ and $\pi'$, used for a stable reference in the training process, follow the soft update $\bm{\theta}' \leftarrow \rho\bm{\theta} + (1-\rho)\bm{\theta}'$ with $\rho\in(0,1)$. TD3 also applies a policy-delay update for stability \cite{fujimoto_AddressingFunctionApproximation_2018}.

In Multi-Agent Reinforcement Learning (MARL), each agent typically observes $\bm{o}^{\langle\tau\rangle} \in \mathcal{O}$ (a partial view of $\bm{s}^{\langle\tau\rangle}$), instead of $\bm{s}^{\langle\tau\rangle}$.

\section{Proposed Method}
\label{sec:alg}
By deploying RL agents at both the server and client levels, pFedMARL achieves improved robustness, fairness, and overall performance while generating personalized models aligned with each client's unique data distribution.
We formulate the FL aggregation as a MARL problem, summarized in Algorithm \ref{alg:rl}:

\begin{figure}[tb]
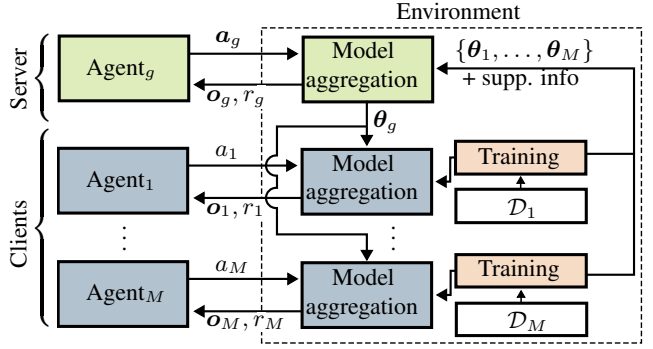

    \centering
    \vspace{1cm}
    \def\svgwidth{0.9\linewidth}
    \include{fig/system/system}
    \vspace{-0.5cm}
    \caption{Overview of the proposed pFedMARL framework, where the server holds one type of agent (green) and every client incorporates another type of agent locally (blue). They return action $\bm{a}$ for model aggregation based on observation $\bm{o}$ and update $\theta^\pi$ regarding reward $\bm{r}$ on the device. The environment illustrates the usual interaction separation in the DRL paradigm.}
    \label{fig:system}
\end{figure}

\begin{algorithm}[tb]
	\SetAlgoLined
	\KwIn{Server $g$, clients $i \in \mathcal{M}$ with $\mathcal{D}_i$, model $\bm{\theta}^{\left<0\right>}$, communication rounds $\tau_\mathit{max}$}
    \For{$\tau \in  \{1, \dots, {\tau}_\mathit{max}$\}}{
        $\bm{a}_{g}^{\left<\tau\right>} \leftarrow \mathrm{Softmax} (\pi_g(\bm{o}_g^{\left<\tau\right>}) + \bm{\varepsilon})$ \\
        $\bm{\theta}_g^{\left<\tau\right>} \leftarrow \sum_{i \in \mathcal{M}} a_{g,i}^{\left<\tau\right>} \bm{\theta}^{\left<\tau-1\right>}_{i}$ \textit{global model update} (\ref{eq:agg_server}) \\

        $r_g^{\left<\tau\right>} = - \log \mathcal{L}(\bm{\theta}_g^{\left<\tau\right>}; \mathcal{D}_{\text{val}})$ \textit{reward (\ref{eq:reward})}\\

        \For {$i \in \mathcal{M}$ \textit{clients in parallel}}{

            $a_{i}^{\left<\tau\right>} \leftarrow \pi_i(\bm{o}_i^{\left<\tau\right>}) + \bm{\varepsilon}$ \\


            $\bm{\theta}_i^{\left<\tau\right>} \leftarrow \mathrm{Adam} (\bm{\theta}_i^{\left<\tau-1\right>}; \bm{\theta}_g^{\left<\tau\right>}, \mathcal{D}_i)$ \textit{with reg. (\ref{eq:client_agg})}\\

            $r_i^{\left<\tau\right>} = - \log \mathcal{L}(\bm{\theta}_i^{\left<\tau\right>}; \mathcal{D}_{\text{val},i})$ \textit{reward (\ref{eq:reward})}\\

            $\bm{o}_i^{\left<\tau+1\right>} \leftarrow (\bm{l}_i, c_i, d_i, \tau)$ \\
        }
        $\bm{o}_g^{\left<\tau+1\right>} \leftarrow (\bm{l}, \bm{c}, \bm{d}, \bm{n}, \tau)$ \\
        \For {$g$ and $i \in \mathcal{M}$ \textit{in parallel}}{
            Store $(\bm{s}^{\left<\tau\right>}, \bm{a}^{\left<\tau\right>}, r^{\left<\tau\right>}, \bm{s}^{\left<\tau+1\right>})$ in $\mathcal{B}$ \\
            Sample transition $\bm{b}$ batch from buffer $\mathcal{B}$ \\
            Update $\bm{\theta}^\pi$, $\bm{\theta}_k^{Q}$, \textit{see  Fig. \ref{fig:agent}} \\ 
            Soft update $\bm{\theta}^{\pi'}$, $\bm{\theta}_k^{Q'}$ via $\bm{\theta}' \leftarrow \rho \bm{\theta} + (1-\rho) \bm{\theta}'$
        }
    }
	\caption{pFedMARL Loop}
	\label{alg:rl}
\end{algorithm}

\textbf{Environment:}
The environment is represented by an FL system with one server and $M$ clients $i \in \mathcal{M} = \{1,\dots,M\}$. Each client $i$ holds a local dataset $\mathcal{D}_i$ and a model $\bm{\theta}_i$. In each communication round $\tau$,  a single RL step proceeds as follows:
1) The server computes observation $\bm{o}_g^{\left<\tau\right>}$ and queries its RL agent.
2) The agent returns a raw action vector $\bm{a}_g^*$, which is normalized via softmax to obtain the impact factors $\bm{a}_g^{\left<\tau\right>}$ in (\ref{eq:agg_server}).
3) The global model $\bm{\theta}_g^{\langle\tau\rangle}$ is broadcasted to all clients.
4) Each client computes its observation $\bm{o}_i^{\left<\tau\right>}$ and sends it to its local agent.
5) The agent outputs an impact factor $a_i^{\left<\tau\right>}$ to mix the global with the locally trained model (\ref{eq:client_agg}).
6) Clients train their new local models $\bm{\theta}_i^{\langle\tau\rangle}$ with global model regularization as in (\ref{eq:agg_server}) and send them, along with the observations, to the server.
7) The next round $\tau+1$ begins.

\textbf{Agent:}
We propose a MARL approach inspired by TD3 \cite{fujimoto_AddressingFunctionApproximation_2018}, with two types of agents: a server agent and client agents. The server's main policy network $\pi(\bm{s})$ dynamically adjusts impact factors for creating a robust global model, while  each client policy balances the global and local models for personalization. Both agents follow the same TD3-inspired training process (Fig. \ref{fig:agent}), using off-policy updates with experience replay buffers $\mathcal{B}$, where transition vectors $\bm{b}$ are saved. They  differ only in the action $\mathcal{A}$ and observation space $\mathcal{O}$.

\textbf{State and Observation:}
The server agent's observation $\bm{o}_g^{\left<\tau\right>}$ consists of four ($M$)-dimensional vectors and the current round index $\tau$, including combined client validation losses $\mathcal{L}(\bm{\theta}_i^{\left<\tau\right>}; \mathcal{D}_{\text{val},i})$ ($\bm{l}^{\left<\tau\right>}$) as described in Section \ref{sec:network}, cosine similarities $\bm{c}^{\left<\tau\right>}$ between client $\Delta\bm{\theta}_i^{\left<\tau\right>}$ and global model updates $\Delta\bm{\theta}_g^{\left<\tau\right>}$, $l^2$-norm distances $\bm{d}^{\left<\tau\right>}$ between global $\bm{\theta}_g^{\left<\tau\right>}$ and client models $\Delta\bm{\theta}_i^{\left<\tau\right>}$, and the number $\bm{n}^{\left<\tau\right>}$ of mini-batches per client. The client agent's observation $\bm{o}_i^{\left<\tau\right>}$ consists of seven single values: cosine similarity, $l^2$-norm distance, and round $\tau$ (as above), the local and global models' separate reconstruction losses $\mathcal{L}_{\text{MSE}}(\bm{\theta}_{i/g}^{\left<\tau\right>}; \mathcal{D}_{\text{val},i})$ and their classification F1-scores $\mathcal{L}_{\text{F1}}(\bm{\theta}_{i/g}^{\left<\tau\right>}; \mathcal{D}_{\text{val},i})$ on local data, combined to $\bm{l}_i^{\left<\tau\right>}$. 

\textbf{Action:}
The server agent's main policy outputs a vector $\bm{a}_g^* = \pi(\bm{o_g^{\left<\tau\right>}}) + \bm{\varepsilon},  \bm{a}_g^{\left<*\right>} \in [-1,1]^M$ with Gaussian exploration noise, converted via softmax into impact factors for aggregation including batch-norm statistics, see (\ref{eq:agg_server}): $\bm{a}_{g}^{\left<\tau\right>} = \text{Softmax}(\bm{a}^{\left<*\right>}_{g})$.
The client agents output a scalar $a_i$ clipped to $[0,1]$ with applied exploration noise, determining the proportion between global regularization and personalized model in the training loss  $\mathcal{L}^{\text{pFedMARL}}_i(\bm{\theta}_i; \bm{\theta}_g^{\left<\tau\right>}, \mathcal{D}_i)$,
\begin{equation}
\label{eq:client_agg}
\mathcal{L}^{\text{pFedMARL}}_i
= \frac{1}{|\mathcal{D}_i|}\sum_{(\bm{x},\bm{y})\in\mathcal{D}_i}
\ell\!\left(f_{\bm{\theta}_i}(\bm{x}),\bm{y}\right)
+ \frac{a_i}{2}\,\left\|\bm{\theta}_i - \bm{\theta}_g^{\left<\tau\right>}\right\|_2^2 \, ,
\end{equation}
where $\ell\!\left(f_{\bm{\theta}_i}(\bm{x}),\bm{y}\right)$ is an arbitrary model loss.

\textbf{Reward:}
The reward incentivizes the validation losses without regularization after each aggregation step as
\begin{equation}
    \label{eq:reward}
    r^{\left<\tau\right>} = - \log \mathcal{L}(\bm{\theta}^{\left<\tau\right>}; \mathcal{D}_{\text{val}}),
\end{equation}
implicitly penalizing performance degradation and compensating for high initial improvements.
Furthermore, due to the same objective, clients and server collaboratively optimize a shared global model that performs well for all clients' data.

\section{Experiments}
\label{sec:experiments}
\begin{figure}[tb]
    \centering
    \resizebox{\linewidth}{!}{%
        \def\svgwidth{1.0\linewidth}
        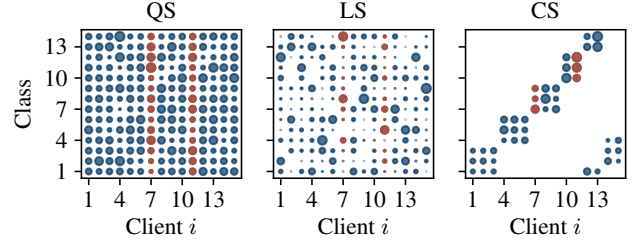
    }
    \vspace{-0.7cm}
    \caption{Visualization of three data partitioning strategies (QS: low quantity \& label skew, LS: high quantity \& label skew, CS: cluster skew) across 14 clients, where the circle size indicates the number of samples per client and the red color adversarial clients.}
    \label{fig:non-iid}
\end{figure}

We evaluate our method on a semi-supervised federated learning task for training a small audio spectrogram transformer (AST) regarding patch reconstruction and classification accuracy. The goal is to train an AST that is effective for data typical for a client's environment while maintaining robust performance on unseen data.

\subsection{Network architecture and training procedure}
\label{sec:network}
We employ a compact, fully connected vision transformer \cite{gong_SSASTSelfSupervisedAudio_2022,baade_MAEASTMaskedAutoencoding_2022} with approximately 450k parameters, suitable for federated scenarios. The model is optimized to minimize the mean squared error (MSE) between patch-masked original and reconstructed log-mel band energy audio features in an unsupervised fashion $\ell_{\text{recon}}\!\left(f_{\bm{\theta}_i}(\bm{x}),\bm{x}\right)$, and the negative log-likelihood loss of a fully-connected classification head attached to the latent space $\ell_{\text{class}}\!\left(f_{\bm{\theta}_i}(\bm{x}),\bm{y}\right)$,  combined to a single loss $\ell = \ell_{\text{recon}} + 2.0\ \ell_{\text{class}}$.

Our TD3 agents are implemented as fully-connected networks with two layers each for the policy ($\pi$) and critic ($Q$), and with 256 units per hidden layer and tanh activation functions. Training uses a batch size of 8, limited to 64 batches per epoch, learning rates of $1\times10^{-2}$ and $1\times10^{-4}$ for policy and critic respectively, discount factor $\gamma=0.80$, soft-update rate $\rho=0.99$, and policy delay of 4 epochs. The Gaussian exploration noise covariance parameter ($\sigma^2$) linearly decreases from 0.40 to 0.05 across 80 epochs. A replay buffer \cite{schaul_PrioritizedExperienceReplay_2016} prioritizing samples with higher TD error is used with parameters $\alpha=0.7$ and $\beta=0.5$.

\subsection{Dataset}
\label{sec:data}
We use $10 \%$ of the DCASE challenge Task 2 development dataset \cite{Dohi2022,Harada2021}, containing single-channel laboratory recordings of machine sounds mixed with environmental noise. It includes audio signals from 14 machine classes (e.g., ToyCar, Fan, Gearbox) sampled at $\SI{16}{\kilo\hertz}$, with audio durations of $\SIrange{6}{18}{\s}$. Each class has 990 normal training clips from the source domain, 10 from the target domain, and 200 test clips (100 normal, 100 anomalous) mixed from both domains, mimicking realistic data drift.

\subsection{Experimental Procedure}

We benchmark pFedMARL against FedAvg \cite{mcmahan_CommunicationEfficientLearningDeep_2017}, Ditto \cite{li_DittoFairRobust_2021} with the optimal tested $\lambda=0.5$, local-only training, and an oracle baseline trained centrally on the raw data of all clients for $\tau_{\text{max}}=100$ FL rounds. Our evaluation compares reconstruction MSE and classification F1-score performance, standard deviation (std) across clients' metrics, and $l^2$-norm between the clients and global model for: client models on local test data (L), the global server model on all client test sets (global data) (S), and client models on global test data (G). The centralized baseline is always tested on global data.

The dataset is divided between $M=15$ clients to simulate three non-IID scenarios: low quantity and label skew (QS, $\alpha=5.0$), high quantity and label skew (LS, $\alpha=0.4$), and cluster skew (CS) with low label imbalance (manual assignment, $\alpha=2.0$). Class distribution probabilities per client are set via a normalized Dirichlet distribution with parameter $\alpha$, enabling realistic data imbalance and variability as illustrated in Figure \ref{fig:non-iid}.

To evaluate the robustness against adversarial clients, we select two clients that apply additive Gaussian noise with $\sigma^2=0.5$ to the transmitted model updates $\Delta \bm{\theta}_i$ while the client's local model remains unchanged, simulating a corrupted transmission or malicious weight injection. 
In the observation $\bm{o}^{\left< \tau \right>}$ only $c_i^{\left< \tau \right>}$ and $d_i^{\left< \tau \right>}$ of the adversarial clients are calculated on corrupted updates, while the losses are based on the non-corrupted models.
This does not apply to the baseline and local training conditions, whose performance will therefore not be degraded by adversarial clients.

\section{Experimental results}
\label{sec:result}
\begin{table}[tb]%
    \caption{Comparison of the FL techniques for the data partitions, shown in Figure \ref{fig:non-iid}. Mean client performance and standard deviation across clients on own local test set (L) and all client test sets (G), and mean $l^2$-norm between client and global model at the end of $\tau_{\text{max}}$.}
    \label{tab:final_label_skew}
    \centering
    \small
    \begin{adjustbox}{max width=1\columnwidth}

\begin{tabular}{ll|cc|cc|cc||cc}
\toprule
 & Algorithm & \multicolumn{2}{c}{pFedMARL} & \multicolumn{2}{c}{Ditto} & \multicolumn{2}{c}{FedAvg} & \multicolumn{2}{c}{Local} \\
 & Data & L & G & L & G & L & G & L & G \\
\midrule
\multirow[c]{5}{*}{QS} & MSE Mean $\downarrow$ & \bfseries 0.10 & \bfseries 0.11 & 0.17 & 0.17 & 1.17 & 1.17 & 0.10 & 0.10 \\
 & MSE Std $\downarrow$ & 0.00 & 0.01 & 0.01 & 0.01 & 0.02 & 0.00 & 0.01 & 0.01 \\
 & F1 Mean $\uparrow$ & \bfseries 0.77 & \bfseries 0.73 & 0.75 & 0.71 & 0.17 & 0.17 & 0.84 & 0.80 \\
 & F1 Std $\downarrow$ & 0.05 & 0.02 & 0.04 & 0.01 & 0.03 & 0.00 & 0.03 & 0.02 \\
 & L2-Norm $\downarrow$ & \multicolumn{2}{c|}{3.32} & \multicolumn{2}{c|}{4.52} & \multicolumn{2}{c||}{2.39} & \multicolumn{2}{c}{3.25} \\
\midrule
\multirow[c]{5}{*}{LS} & MSE Mean $\downarrow$ & \bfseries 0.10 & \bfseries 0.11 & 0.17 & 0.18 & 0.96 & 0.96 & 0.08 & 0.11 \\
 & MSE Std $\downarrow$ & 0.02 & 0.02 & 0.02 & 0.03 & 0.08 & 0.00 & 0.02 & 0.03 \\
 & F1 Mean $\uparrow$ & \bfseries 0.87 & \bfseries 0.60 & 0.69 & 0.34 & 0.13 & 0.13 & 0.92 & 0.59 \\
 & F1 Std $\downarrow$ & 0.05 & 0.08 & 0.07 & 0.06 & 0.09 & 0.00 & 0.04 & 0.08 \\
 & L2-Norm $\downarrow$ & \multicolumn{2}{c|}{3.31} & \multicolumn{2}{c|}{6.17} & \multicolumn{2}{c||}{2.35} & \multicolumn{2}{c}{3.22} \\
\midrule
\multirow[c]{5}{*}{CS} & MSE Mean & \bfseries 0.06 & \bfseries 0.12 & 0.15 & 0.19 & 1.25 & 1.25 & 0.03 & 0.07 \\
 & MSE Std $\downarrow$ & 0.03 & 0.07 & 0.05 & 0.07 & 0.10 & 0.00 & 0.03 & 0.06 \\
 & F1 Mean $\uparrow$ & \bfseries 0.96 & \bfseries 0.21 & 0.91 & 0.19 & 0.02 & 0.02 & 0.98 & 0.21 \\
 & F1 Std $\downarrow$ & 0.04 & 0.01 & 0.09 & 0.02 & 0.04 & 0.00 & 0.04 & 0.01 \\
 & L2-Norm $\downarrow$ & \multicolumn{2}{c|}{3.32} & \multicolumn{2}{c|}{4.69} & \multicolumn{2}{c||}{2.40} & \multicolumn{2}{c}{3.19} \\
\bottomrule
\end{tabular}

    \end{adjustbox}
\end{table}
\begin{table}[tb]%
    \caption{Comparison of the FL techniques server model on all client test sets for the three non-IID scenarios at the end of $\tau_{\text{max}}$.}
    \label{tab:final_label_skew_server}
    \centering
    \small
\begin{tabular}{l|ccc|ccc}
\toprule
Metric & \multicolumn{3}{c}{F1 Mean $\uparrow$} & \multicolumn{3}{c}{MSE Mean $\downarrow$} \\
Scenario & CS & LS & QS & CS & LS & QS \\
\midrule
pFedMARL & \textbf{0.22} & \textbf{0.38} & \textbf{0.61} & \textbf{0.18} & \textbf{0.24} & \textbf{0.24} \\
Ditto & 0.11 & 0.07 & 0.22 & 2.07 & 79.19 & 2.17 \\
FedAvg & 0.03 & 0.12 & 0.17 & 1.34 & 1.02 & 1.23 \\
\midrule
Baseline & \multicolumn{3}{c|}{0.97} & \multicolumn{3}{c}{0.01} \\
\bottomrule
\end{tabular}

\end{table}

\begin{figure}[tb]
    \centering
    \resizebox{\linewidth}{!}{%
        \def\svgwidth{1.1\linewidth}
        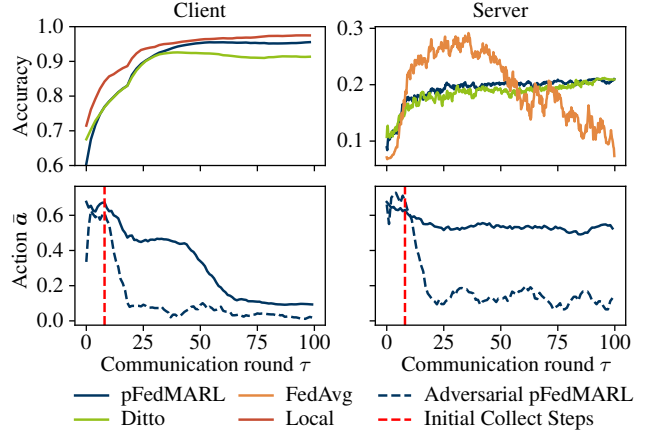
    }
    \vspace{-0.7cm}
    \caption{Mean validation client on local data and server F1-accuracy on all client sets, along with mean actions $a^{\left<\tau\right>}$ separated into adversarial clients and the other clients for the CS non-IID scenario. 
    }
    \label{fig:rounds}
\end{figure}

Table \ref{tab:final_label_skew} summarizes the final performance of each client and algorithm on their local test sets (L) and the test sets of all clients (G) for the three non-IID scenarios. 
Local-only training excels in personalization and also performs well in reconstructing  global data, where data diversity is less critical, but falters in classifying unseen classes.
With adversarial clients, pFedMARL consistently surpasses FedAvg and Ditto in both MSE and F1-score for the local and global evaluation. 
As pFedMARL optimizes for local performance, it converges towards local-only training, nearly matching its global performance yet falling short on local data due to the adversarial influence.
Across scenarios, we observe the expected personalization/gen\-er\-ali\-zation trade-off, where gains in L coincide with losses in G. 

Server models shown in Table \ref{tab:final_label_skew_server} underperform compared to client models on global data (G) because adversarial updates corrupt the aggregated weights more. pFedMARL does mitigate this by down-weighting adversarial clients but cannot exclude them entirely.
The baseline is unaffected by data heterogeneity or adversarial weight injection due to the availability of all data at a centralized server and serves as the overall upper bound.

Figure \ref{fig:rounds} illustrates these dynamics. After the initial collection phase, the average contribution of benign clients to the global model stabilizes near 0.5, whereas adversarial influence drops to $\approx 0.1$. This rapid suppression coincides with a steady rise in server accuracy after the initial transition. On the client side, adversarial participants choose almost no global regularization, while benign clients initially leverage the global model (actions near the Ditto optimum of $\lambda \approx 0.5$) to exploit shared knowledge and, after $\approx 50$ rounds, reduce reliance on the server model to personalize further. Correspondingly, pFedMARL’s local F1-accuracy tracks Ditto until about round 50 and then increases beyond it, approaching local training.

\section{Conclusions}
\label{sec:conclusions}
In this paper, we introduce pFedMARL\footnote{ Code available under: \href{https://github.com/NexuFed/pFedMARL}{github.com/NexuFed/pFedMARL}}
, an adaptive FL framework utilizing multi-agent reinforcement learning to address non-IID challenges and adversarial client behavior effectively. By employing TD3-based agents on both server and client sides, our approach achieves a robust global model while ensuring personalized client models without pre-training the agents. Empirical evaluations in an audio spectrogram transformer training task across various data skew scenarios with adversaries demonstrate that pFedMARL may outperform traditional aggregation methods such as FedAvg, Ditto and local-only training.
Particularly, pFedMARL shows consistent performance in adversarial scenarios and matches Ditto in non-adversarial settings, delivering high accuracy with improved fairness among clients and robustness while bypassing Ditto's need for a second local training. 
Thus, our results highlight the potential of collaborative multi-agent reinforcement learning to enhance FL practices in real-world applications. Future research will explore the extension of pFedMARL to other data modalities and evaluate scalability in federations with significantly more clients.

\clearpage

\section{Acknowledgement}
This work has been supported by 
the German Federal Ministry of Research, Technology and Space (BMFTR)
(grant 02L19C200, "HUMAINE") 
and by the German Research Foundation (DFG) (project numbers 429873205 and 549576906).

\bibliographystyle{IEEEbib}
\bibliography{refs}


\end{document}